\documentclass{article}

\usepackage{microtype}
\usepackage{graphicx}
\usepackage{subfigure}
\usepackage{booktabs} 
\usepackage{caption}
\usepackage{enumitem}
\usepackage{hyperref}

\usepackage[accepted]{icml2025}

\usepackage{amsmath}
\usepackage{amssymb}
\usepackage{mathtools}
\usepackage{amsthm}
\usepackage{comment}
\usepackage[capitalize,noabbrev]{cleveref}

\theoremstyle{plain}

\theoremstyle{definition}

\theoremstyle{remark}

\usepackage[textsize=tiny]{todonotes}

\begin{document}

\twocolumn[
\icmltitle{SOLAR: Scalable Optimization of Large-scale Architecture for Reasoning}



\icmlsetsymbol{equal}{*}

\begin{icmlauthorlist}
    \icmlauthor{Chen Li}{equal,cmu}
    \icmlauthor{Yinyi Luo}{equal,cmu}
    \icmlauthor{Anudeepsekhar Bolimera}{cmu}
    \icmlauthor{Uzair Ahmed}{indep}
    \icmlauthor{Shri Kiran Srinivasan}{indep}
    \icmlauthor{Hrishikesh Gokhale}{cmu}
    \icmlauthor{Marios Savvides}{cmu}
\end{icmlauthorlist}

\icmlaffiliation{cmu}{Carnegie Mellon University, Pittsburgh, PA, USA}
\icmlaffiliation{indep}{Independent Researcher}

\icmlcorrespondingauthor{Chen Li}{chenli4@andrew.cmu.edu}
\icmlcorrespondingauthor{Yinyi Luo}{yinyil@andrew.cmu.edu}
\icmlcorrespondingauthor{Anudeep Bolimera}{abolimer@andrew.cmu.edu}
\icmlcorrespondingauthor{Uzair Ahmed}{uzair.cmu@gmail.com}
\icmlcorrespondingauthor{Shri Kiran Srinivasan}{
shrikiran114@gmail.com}
\icmlcorrespondingauthor{Hrishikesh Gokhale}{
hgokhale@andrew.cmu.edu}
\icmlcorrespondingauthor{Marios Savvides}{marioss@andrew.cmu.edu}

\icmlkeywords{Machine Learning, Topological Reasoning, LLMs} 

\icmlkeywords{Machine Learning, ICML}

\vskip 0.3in
]



\printAffiliationsAndNotice{\icmlEqualContribution} 

\begin{abstract}
Large Language Models excel in reasoning yet often rely on Chain-of-Thought prompts, limiting performance on tasks demanding more nuanced topological structures. We present \textbf{SOLAR} (Scalable Optimization of Large-scale Architecture for Reasoning), a framework that dynamically optimizes Chain-of-Thought (CoT), Tree-of-Thought (ToT), and Graph-of-Thought (GoT) topologies to boost accuracy and efficiency.

Our Topological-Annotation-Generation (TAG) system automates dataset creation, annotation, and difficulty segmentation, leading to stronger post training and test-time performance. We also propose Topological-Scaling, a curriculum-learning-based approach that adaptively combines post training and inference scaling to each task. On MATH and GSM8K, SOLAR delivers notable gains: \textbf{+5\%} accuracy with Topological Tuning, \textbf{+9\%} with Topological Rewarding, and \textbf{+10.02\%} with Hybrid Scaling, while reducing response length by over \textbf{5\%}, lowering inference latency.

To further enhance efficiency, we introduce a multi-task Topological Reward Model (\textbf{M-TRM}) that selects both the optimal reasoning topology and final answer in a single pass, eliminating multiple single-task TRMs. Remarkably, M-TRM also surpasses all single-task TRMs, improving accuracy by \textbf{+10\%} and rank correlation by \textbf{+9\%}.

Overall, SOLAR establishes a new benchmark for scalable, high-precision LLM reasoning and introduces a fully automated, dynamic topology competition mechanism.
\end{abstract}

\vspace{-0.9cm}
\section{Introduction}
\label{submission}
Large Language Models (LLMs) excel at complex reasoning but typically rely on sequential Chain-of-Thought (CoT) prompts. Many real-world tasks, however, require more nuanced topological strategies (e.g., trees, graphs). We introduce \textbf{SOLAR} (Scalable Optimization of Large-scale Architecture for Reasoning), a framework that dynamically selects the optimal topology for each problem, thereby enhancing LLM performance.

\subsection{Observations on LLM Reasoning}
Our systematic evaluations reveal:
\begin{itemize}
    \item LLMs default to Chain-of-Thought (CoT) reasoning and rarely generate more sophisticated structures like Tree-of-Thought (ToT) or Graph-of-Thought (GoT) without explicit prompting.
    \item Complex tasks (e.g., TSP, multi-stage robotics manipulation) benefit from alternative topologies, surpassing default Chain-of-Thought (CoT) performance.
\end{itemize}

\subsection{Our Approach}
\begin{figure*}[h!]
    \centering
    \includegraphics[width=0.6\textwidth]{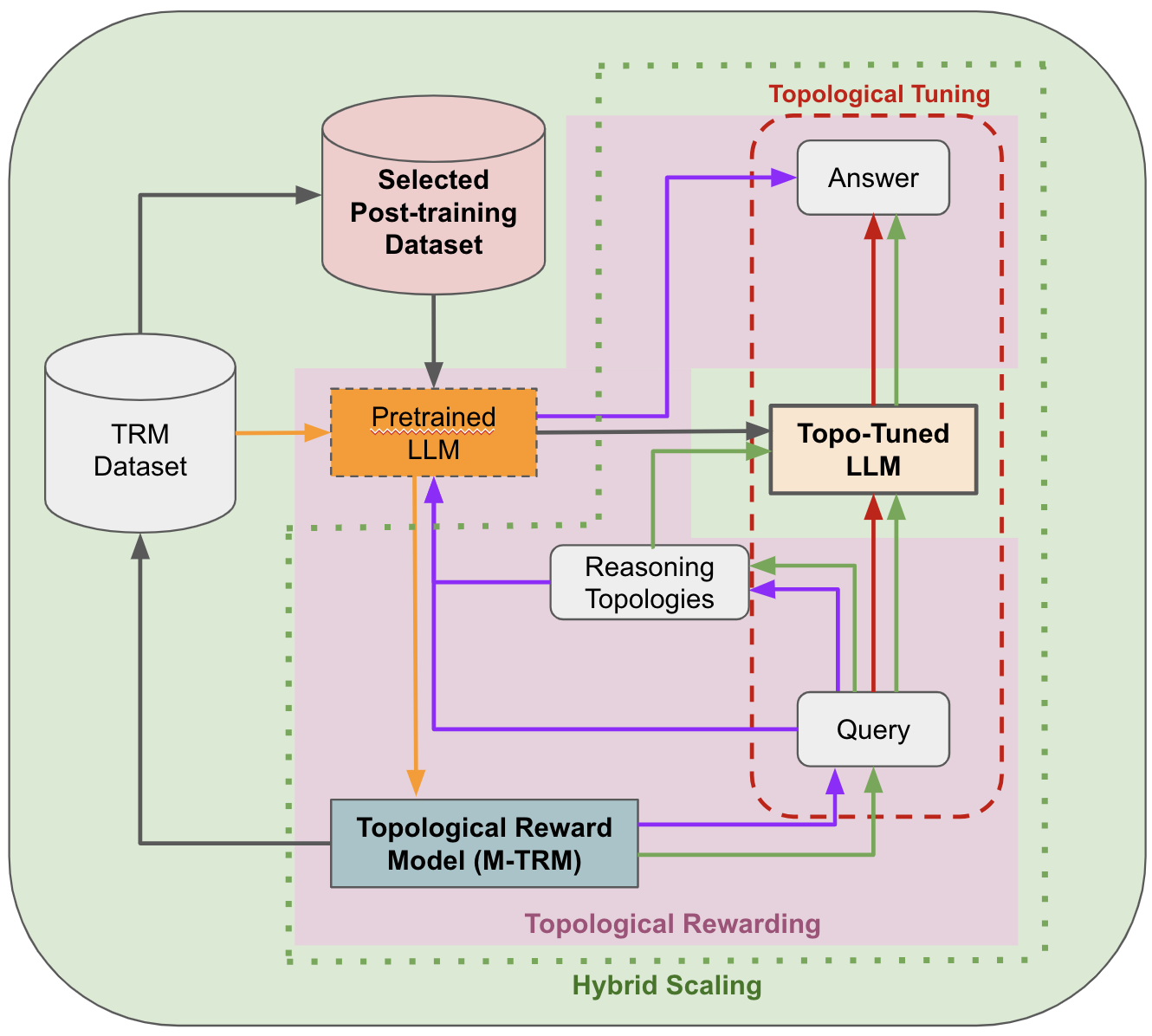}
    \caption{SOLAR Architecture}
    \label{fig:pipeline.png}
\end{figure*}

We hypothesize that distinct reasoning problems demand different topologies for higher accuracy. To validate this and boost reasoning capabilities, we propose a multi-stage pipeline:

\paragraph{Evaluation Pipeline}
We compare CoT, ToT, and GoT on MATH and GSM8K across diverse model sizes. Three insights emerge:
\begin{itemize}
    \item Different tasks favor different topologies, each offering unique accuracy advantages, shown from Win Rate characterization.
    \item ToT and GoT maintain CoT-level accuracy, despite being less frequently generated.
    \item This pattern holds for both smaller and large-scale state-of-the-art models.
\end{itemize}

\paragraph{Synthetic Topological Data Infrastructure} 
We automate dataset creation and annotation for multiple topologies, enabling problem difficulty segmentation based on multi-dimensional reasoning performance data. Similar methods (e.g., \citealt{ding2024easy2hardbenchstandardizeddifficultylabels}) do not consider topological factors, potentially missing critical reasoning attributes.

\paragraph{Topological-Scaling Framework} 
We introduce a competitive selection process that identifies the best reasoning topology and final answer at test time, unifying post training and inference scaling strategies:
\begin{itemize}
    \item \textbf{Topological Tuning}: Supervised Fine-Tuning \citep{zhang2024instructiontuninglargelanguage} to generate optimal topology policies (+5\% accuracy). For complex tasks (e.g., MATH), it also reduces token length by 5\%.
    \item \textbf{Topological Rewarding}: Our inference scaling method leveraging a multi-task topological reward model (M-TRM) to select the optimal topology and answer, achieving a $+9\%$ accuracy gain at the cost of increased latency. A single forward pass determines both the topology and the final answer.
    \item \textbf{Hybrid Scaling}: Integrating training-time and inference-time scaling to maximize performance, achieving a $+10.02\%$ accuracy gain at the cost of increased computation.
\end{itemize}

Figure~\ref{fig:pipeline.png} outlines our overall architecture. We conduct extensive experiments to characterize the trade-offs among efficiency, computational cost, and generation accuracy.

\subsection{Contributions}
\begin{itemize}
    \item \textbf{Topological Reasoning Characterization}: We systematically show that different tasks require distinct topologies, a phenomenon validated across various models and datasets.
    \item \textbf{Topological-Annotation-Generation (TAG)}: An automated system to build and annotate large-scale topological datasets, including difficulty segmentation, facilitating robust post training.
    \item \textbf{Hierarchical Topological-Scaling Framework}: A unified mechanism combining post training and inference scaling optimizations, significantly boosting performance while allowing flexible trade-offs between accuracy and efficiency.
\end{itemize}

Our results demonstrate substantial gains on MATH and GSM8K, underscoring how curriculum learning-based topological scaling effectively enhances LLM reasoning capabilities with retained efficiency.

\section{Related Work}

\subsection{Reward Models and Scaling Laws}
Reward Models (RMs) guide both training and inference in Large Language Models (LLMs) by providing numerical feedback. They primarily fall into two categories: Outcome Reward Models (ORM), which evaluate final outputs (as in RLHF \citep{ouyang2022traininglanguagemodelsfollow} and RLAIF \citep{bai2022traininghelpfulharmlessassistant}), and Process Reward Models (PRM), which score intermediate reasoning steps \citep{lightman2023letsverifystepstep}. While training-scale approaches focus on expanding model size or dataset volume \citep{shuai2024scalinglawlanguagemodels}, inference scaling adjusts reasoning depth dynamically at test time \citep{wu2024inferencescalinglawsempirical}. Leveraging step-level rewards and iterative refinements can significantly enhance multi-step reasoning \citep{zeng2024bstarmonitoringbalancingexploration}.

\subsection{Reinforcement Learning in LLMs Reasoning}
Recent advancements in RLHF \citep{stiennon2022learningsummarizehumanfeedback, ouyang2022traininglanguagemodelsfollow} and RLAIF \citep{bai2022constitutionalaiharmlessnessai} highlight the effectiveness of reward-guided optimization. Algorithms generally fall into two categories: (1) Reward-Based Methods, such as PPO \citep{schulman2017proximalpolicyoptimizationalgorithms}, RPO \citep{yin2024relativepreferenceoptimizationenhancing}, and GRPO \citep{shao2024deepseekmathpushinglimitsmathematical}, and (2) Reward-Free Methods, including DPO \citep{rafailov2024directpreferenceoptimizationlanguage}, SIMPO \citep{meng2024simposimplepreferenceoptimization}, and ORPO \citep{hong2024orpomonolithicpreferenceoptimization}. Extending these approaches to multi-topology reasoning can further enhance both accuracy and interpretability.

\subsection{Advances in Topological Reasoning}
While Chain-of-Thought (CoT) \citep{wei2023chainofthoughtpromptingelicitsreasoning} is widely used, Tree-of-Thought (ToT) \citep{yao2023treethoughtsdeliberateproblem} and Graph-of-Thought (GoT) \citep{Besta_2024} have emerged to tackle more complex tasks, such as TSP and multi-stage decision making. Existing methods often fix reasoning topology by default, but our work dynamically learns which topology best suits each problem, enabling more flexible and accurate reasoning.

\subsection{Curriculum Learning for Structured Reasoning}
Curriculum learning \citep{bengio2009curriculum} gradually introduces tasks of increasing difficulty. Applications include reverse curriculum RL \citep{xi2024traininglargelanguagemodels}, iterative expert self-training \citep{zhao2024automaticcurriculumexpertiteration}, and problem-solving heuristics \citep{ma2025problemsolvinglogicguidedcurriculum} to refine reasoning. When combined with reward modeling, curriculum strategies can further optimize both training efficiency and inference performance.

Overall, we are the first to systematically integrate multi-topology curriculum learning with both post training and inference scaling paradigms, redefining LLM optimization for complex problem-solving.

\section{Methodology}

\subsection{Hypothesis Validation and Evaluation Methods}

\subsubsection{Observations and Hypothesis} \label{sec:hypo}

We begin by analyzing the reasoning patterns of LLMs when solving mathematical problems. Through systematic evaluation, we observe the following phenomena:

\begin{itemize}
    \item LLMs primarily generate Chain-of-Thought (CoT) reasoning and rarely employ more advanced structures like Tree-of-Thought (ToT) or Graph-of-Thought (GoT).
    \item Problems such as \textit{Data Center Fault Tolerance}, the \textit{Traveling Salesman Problem (TSP)}, and \textit{Multi-Stage Robotic Manipulation} require advanced topological reasoning structures beyond CoT to achieve optimal solutions.
\end{itemize}

Based on these observations, we propose the following two hypotheses to be validated in later sections:

\begin{itemize}
    \item \textbf{Hypothesis 1:} Different problems require distinct optimal reasoning topologies that yield the best solutions.
    \item \textbf{Hypothesis 2:} Solving problems with optimal topological reasoning structures can significantly enhance generation accuracy.
\end{itemize}

\subsubsection{Validating Hypothesis 1: Topological Annotation and Evaluation} \label{sec:hypo}

To validate \textbf{Hypothesis 1}, we designed and implemented an automated data generation and annotation system, the Topological-Annotation-Generation (TAG) System (detailed in Section~\ref{sec:data_infra}). This system constructs a synthetic dataset where each sample consists of: (1) a problem statement paired with a group of generated responses, (2) multiple reasoning topologies, including CoT, ToT, and GoT, and (3) a hierarchical labeling system annotated automatically.

Specifically, this hierarchical labeling system is illustrated as below. Each sample in the dataset is automatically annotated with two labels:

\begin{itemize}
    \item \textbf{Topo Label:} A continuous value in the range \([0,1]\), representing the probability that a given topology produces the correct answer for a question.
    \item \textbf{Hard Label:} A binary value \(\{0,1\}\), indicating whether the generated answer is correct.
\end{itemize}

With these labels, we evaluate each reasoning topology by defining the following two metrics:

\begin{itemize}
    \item \textbf{Accuracy:} The proportion of correct answers generated using each topology.
    \item \textbf{Win Rate:} The likelihood of each topology being the best-performing structure across all questions.
\end{itemize}

\paragraph{Win Rate Calculation}

The \textbf{Win Rate} of a topology \( T \in \{CoT, ToT, GoT\} \) is defined as:

\begin{scriptsize}
\begin{equation}
    \text{WinRate}(T) = \frac{|\{ q \in Q \mid T = \arg\max_{T' \in \{CoT, ToT, GoT\}} \text{Topo-label}(q, T') \}|}{|Q|}
\end{equation}
\end{scriptsize}

where \( Q \) is the total set of questions, and \( \text{Topo-label}(q, T) \) denotes the topo-label of topology \( T \) for question \( q \). For each question, the topology with the highest topo-label is assigned a win. The win rate for each topology is then computed as the fraction of questions where it was optimal.

Experimental results (detailed in Section~\ref{sec:phenomenon}) confirm that different problems exhibit different optimal topological reasoning structures, a phenomenon agnostic to model size or capacity, thus validating Hypothesis 1.

\subsubsection{Validating Hypothesis 2: Performance Boost With Topological Scaling} \label{sec:hypo2}

To validate \textbf{Hypothesis 2}, we design and implement a hierarchical, adaptively curriculum-learning based framework, Topological Scaling, which harnesses the synergy between post training and inference scaling in a multi-topological reasoning space. We conduct rigorous ablation studies to evaluate the impact of our approach.

Experimental results (presented in Section~\ref{sec:experiments}) demonstrate significant performance improvements, further supporting the Hypothesis 2. The details of our methodology are illustrated in Section~\ref{sec:training}.

\subsection{Synthetic Topological Data Infrastructure}\label{sec:data_infra}

\subsubsection{Topological-Annotation-Generation System (TAG)}\label{sec:data_infra}

In this section, we outline our approach in automatically annotating the topology reasoning dataset. We begin by introducing the datasets used in our study, followed by a detailed breakdown of data generation and annotation process.

\paragraph{Datasets} This work leverages two datasets: GSM8K \citep{cobbe2021trainingverifierssolvemath} and MATH \citep{hendrycks2021measuringmathematicalproblemsolving}. For training purpose, we split both datasets to training and testing sets. The final constructed synthetic data can be used for both post training purpose and for evaluation purpose. 

\paragraph{Data Generation}
To ensure diversity in reasoning topologies and a balanced distribution of positive and negative samples in our dataset, we utilize both a small-scale model, Qwen2-VL-7B-Instruct \citep{wang2024qwen2vlenhancingvisionlanguagemodels}, and an open-source state-of-the-art reasoning model with hundreds of billions of parameters. These models generated responses across three reasoning topologies—Chain-of-Thought (CoT), Tree-of-Thought (ToT), and Graph-of-Thought (GoT)—with extensive degree of freedom in maximum depth, number of children, and number of neighbors.

\paragraph{Automatic Annotation} 
As described in Section~\ref{sec:hypo}, we assign each problem a Topo Label and each response a Hard Label. We design an automated annotation pipeline for topological reasoning as follows:  

First, using the generation mechanism outlined in the paragraph above, we obtain a diverse set of responses for each question, covering all three reasoning topologies—CoT, ToT, and GoT. We then apply the following annotation process to the generated reasoning data:  

- \textbf{Topo Label} (\(\mathcal{T}_q\)): This problem-specific label reflects how effectively each reasoning topology solves a given problem. For each problem 
\( q \), we compute the accuracy of responses from each topology and assign it as the problem:
  \begin{equation}
      \mathcal{T}_q = \max_{T \in \{\text{CoT, ToT, GoT}\}} \frac{N_{\text{correct}}(q, T)}{N_{\text{total}}(q, T)}
      \label{eq:topo_label}
  \end{equation}

  where \( N_{\text{correct}}(q, T) \) is the number of correct responses using topology \( T \) for question \( q \), and \( N_{\text{total}}(q, T) \) is the total number of responses generated using \( T \). The resulting \(\mathcal{T}_q\) is a continuous value in \([0,1]\).  

- \textbf{Hard Label} (\(\mathcal{H}_a\)): This is a response-specific label which is a variant of a binary Outcome-Reward-Model(ORM) label. Each response \( a \) is assigned a 1 if correct and 0 if incorrect:

  \begin{equation}
      \mathcal{H}_a =
      \begin{cases}
        1, & \text{if } a \text{ is correct} \\
        0, & \text{if } a \text{ is incorrect}
      \end{cases}
      \label{eq:hard_label}
  \end{equation}

These annotations allow us to quantitatively evaluate the performance of different reasoning topologies and assess their impact on problem-solving accuracy. 

\subsubsection{Problems Difficulty Segmentation}\label{sec:diff}

With \textbf{TAG}, we gain an additional advantage: the ability to analyze problems from an entirely new perspective. By examining the distribution of Topo Labels across all three reasoning structures, we can redefine problem difficulty in a multi-dimensional data-driven manner, with considerations from both outcomes and reasoning process, providing nutritious data for downstream post training tasks and a toolkit for further finer-grained research. Specifically, we categorize problems as follows:

\begin{itemize} \item \textbf{Hard}: Problems where all three Topo Labels fall below a specified quantile threshold in their respective distributions.
\item \textbf{Easy}: Problems where all three Topo Labels exceed a specified quantile threshold in their respective distributions.
\item \textbf{Medium}: Problems that do not fall into either the hard or easy categories.
\end{itemize}

\subsection{Topological Scaling for Enhanced Reasoning} \label{sec:training}
\paragraph{Topological Tuning} We perform Supervised Fine-Tuning (SFT) on topological reasoning data carefully selected  by TAG, which is split into train and test sets. Training data is produced through the following three-step process:

\begin{itemize}
    \item \textbf{Diversity Sampling}: To ensure a balanced dataset, we sample the same proportion of data from hard,  easy, and medium problems, respectively, based on the difficulty segmentation defined in Section~\ref{sec:diff}.
    \item \textbf{Correct Answer Filtering}: For finetuning purpose, we keep correct responses only, which have positive Hard Labels.
    \item \textbf{Rejection Sampling (RS)}: Following \citep{grattafiori2024llama3herdmodels, qwen2025qwen25technicalreport}, we apply RS using an in-housed well-trained multi-task topological reward model (M-TRM) to remove spurious samples. The reward model is detailed in the next paragraph.
\end{itemize}

We then train the model using Next Token Prediction \citep{wang2024emu3nexttokenpredictionneed} on this curated dataset. The base model for SFT is Qwen2-VL-7B-Instruct \citep{wang2024qwen2vlenhancingvisionlanguagemodels}, with finetuning performed using LoRA \citep{hu2021loralowrankadaptationlarge} for parameter-efficient adaptation.

This post training strategy is optimized for real-time applications that demand low inference latency and high accuracy. As shown in Section~\ref{sec:fintune}, finetuning the model with diverse topological reasoning data surpasses the baseline, producing shorter yet more accurate responses, ultimately reducing latency.

\paragraph{Topological Rewarding}  
At inference time, we introduce a \textit{Topology Competition Game}, where a base model (with or without finetuning) generates responses using multiple reasoning topologies. Our multi-task reward model (M-TRM) then selects both the optimal topology and the best final answer in a single pass.

This system supports two modes:  
1) \textbf{Inference Scaling Only}, where an unfine-tuned base model relies solely on M-TRM for selection;  
2) \textbf{Hybrid Scaling}, where a fine-tuned base model (e.g., via topological tuning) is combined with inference-time selection in a curriculum-style setup (see next paragraph). Experimental results are reported in Section
~\ref{sec:inf_scale}.
M-TRM is trained using a multi-task objective: Topo Labels (regression loss; Equation~\eqref{eq:mse}) and Hard Labels (pairwise ranking loss; Equation~\eqref{eq:ranking_loss}).

\textbf{Mean Squared Error (MSE) Loss:}

\begin{equation}
\mathcal{L}_{\text{MSE}} = \frac{1}{n} \sum_{i=1}^{n} \left( \hat{y}_i - y_i \right)^2
\label{eq:mse}
\end{equation}

where \( \hat{y}_i \) is the predicted reward score, \( y_i \) is the ground-truth scalar reward, and \( n \) is the number of samples.

\textbf{Pairwise Ranking Loss:}

\begin{equation}
\mathcal{L}_{\text{rank}} = \frac{1}{n} \sum_{i=1}^{n} \log \left( 1 + \exp\left( -\beta \left( \hat{y}_i^+ - \hat{y}_i^- \right) \right) \right)
\label{eq:ranking_loss}
\end{equation}

where \( \hat{y}_i^+ \) and \( \hat{y}_i^- \) are the predicted scores for the preferred and dispreferred completions in the \( i \)-th pair, and \( \beta \) is a scaling hyperparameter (often set to 1).

To evaluate the quality of M-TRM independently from downstream performance, we use Spearman rank correlation~\citep{spearman1904correlation} for the regression task on Topo Labels (Equation~\eqref{eq:spearman}), and Accuracy, defined in Section~\ref{sec:hypo} for the pair-wise ranking task on Hard Labels.

\textbf{Spearman Rank Correlation:}

\begin{equation}
\rho = 1 - \frac{6 \sum_{i=1}^{n} (r_i - \hat{r}_i)^2}{n(n^2 - 1)}
\label{eq:spearman}
\end{equation}

where \( r_i \) and \( \hat{r}_i \) are the ranks of the ground-truth and predicted Topo Labels for the \( i \)-th instance, respectively, and \( n \) is the total number of instances.

\paragraph{Hybrid Scaling} This follows the second usage scenario described above, where the base generation model is a topologically tuned model. This approach seamlessly combines SFT with inference scaling, achieving the highest performance gains. However, it requires increased computation during both training and inference, leading to higher latency. This strategy is best suited for downstream tasks that align with its performance objectives and computational constraints. Experimental results are presented in Section~\ref{sec:inf_scale}.

\section{Experiments} \label{sec:experiments}

\subsection{Experiment Setup} \label{sec:setting}
We evaluate our method on complex mathematical problems from GSM8K and MATH, using TAG to create a topological reasoning dataset with annotated Topo Labels and Hard Labels. We split the data into training and test sets and run all experiments on eight NVIDIA A100 GPUs.

To measure performance, we use \emph{Accuracy} and \emph{Win Rate} (Section~\ref{sec:hypo}), evaluated both per topology and overall. As a baseline, we select Qwen2-VL-7B-Instruct \citep{wang2024qwen2vlenhancingvisionlanguagemodels} in its non-finetuned form due to its strong multi-topology capabilities. In Section~\ref{sec:inf_scale}, we also include Qwen2.5-Math-7b~\citep{yang2024qwen25mathtechnicalreportmathematical} for a broader performance comparison.

The rest of this section proceeds as follows: Section~\ref{sec:phenomenon} tests Hypothesis~1 (Section~\ref{sec:hypo}), and Sections~\ref{sec:fintune}--\ref{sec:inf_scale} address Hypothesis~2 (Section~\ref{sec:hypo2}). Specifically, Section~\ref{sec:fintune} examines Topological Tuning with an ablation study, and Section~\ref{sec:inf_scale} focuses on Topological Rewarding and Hybrid Scaling. Finally, Section~\ref{sec:discuss} discusses our design considerations and the trade-offs of our three proposed strategies.

\subsection{Topological Reasoning Validation} \label{sec:phenomenon}
This section validates Hypothesis~1 (Section~\ref{sec:hypo}) by examining models’ abilities to generate ToT and GoT for 1{,}000 questions, each prompted five times with three reasoning topologies. As shown in Table~\ref{tab:success_rates}, Qwen2.5-Math \citep{yang2024qwen25mathtechnicalreportmathematical} achieves an 11\% success rate, while another leading model reaches 7\%. In contrast, Qwen2-VL-7B-Instruct \citep{wang2024qwen2vlenhancingvisionlanguagemodels} attains 74\%, confirming our choice of base model.


\begin{table}[h]
    \centering
    \begin{tabular}{|l|c|}
        \hline
        \textbf{Model} & \textbf{Success Rate (\%)} \\
        \hline
        Qwen2.5-Math & 11 \\
        Leading Math Model & 7 \\
        Qwen2-VL-7B-Instruct & 74 \\
        \hline
    \end{tabular}
    \caption{Success Rate of Generating Multi-Topo from Different Models}
    \label{tab:success_rates}
\end{table}

We hypothesize that Qwen2-VL’s advantage stems from its exposure to diverse high-dimensional training data, potentially enhancing non-Euclidean representation. Future work will delve deeper into this phenomenon.

\begin{figure}[h!]
    \centering
    \includegraphics[width=0.5\textwidth]{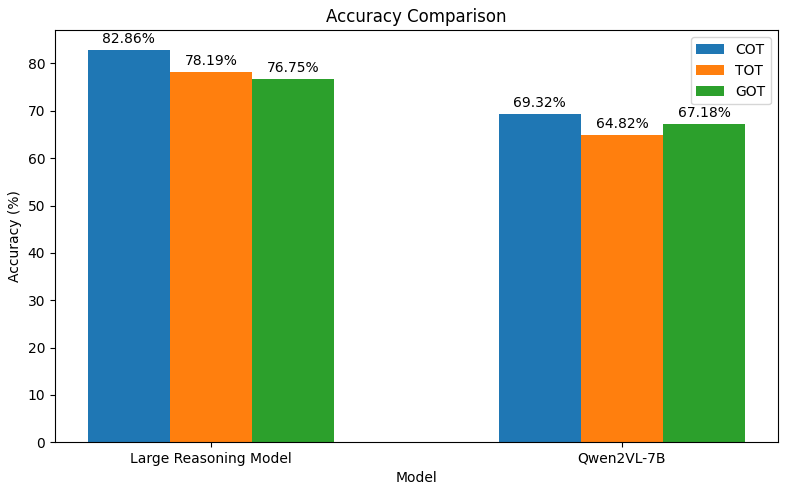}
    \caption{Accuracy comparisons across existing pretrained models reveal that the less frequently generated ToT and GoT topologies perform on par with the default CoT method, indicating that neither ToT nor GoT is lagging behind in performance.}
    \label{fig:acc_com_exisiting models}
\end{figure}

\begin{figure}[h!]
    \centering
    \includegraphics[width=0.5\textwidth]{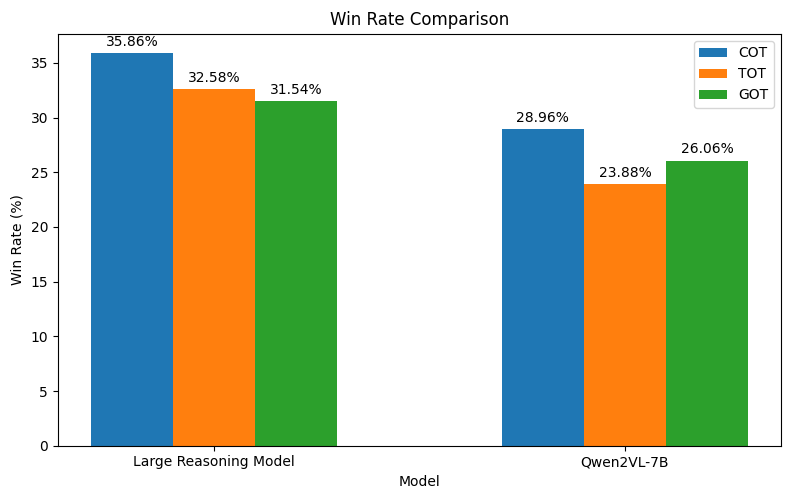}
    \caption{Win Rate comparisons across pretrained models demonstrate that different tasks favor different reasoning topologies, as evidenced by distinct win-rate distributions. This finding underscores the potential to enhance LLM reasoning by explicitly augmenting them with optimal topological strategies.}
    \label{fig:win_rate_exisiting models}
\end{figure}

Figures~\ref{fig:acc_com_exisiting models} and \ref{fig:win_rate_exisiting models} show that although ToT and GoT are generated less frequently, they achieve comparable accuracy to CoT and therefore fall behind in overall performance. Moreover, their Win Rate distributions reveal that different tasks favor different reasoning topologies, indicating the universality and scale-agnostic nature of multi-topology reasoning. These findings confirm Hypothesis~1, with detailed metrics in Table~\ref{tab:accuracy_winrate}.

Sections~\ref{sec:fintune} and \ref{sec:inf_scale} will next validate Hypothesis~2.

\subsection{Topological Tuning Impact} \label{sec:fintune}
\subsubsection{Topological Tuning Results}  \label{sec:topo_fintune}
We finetuned Qwen2VL-7B-Instruct model using training data which is annotated by TAG and curated following a filtering process illustrated in Section~\ref{sec:training}, and then mixed with alpaca dataset \citep{alpaca2023} to prevent catastrophic forgetting. 

\begin{figure}[h!]
    \centering
    \includegraphics[width=0.5\textwidth]{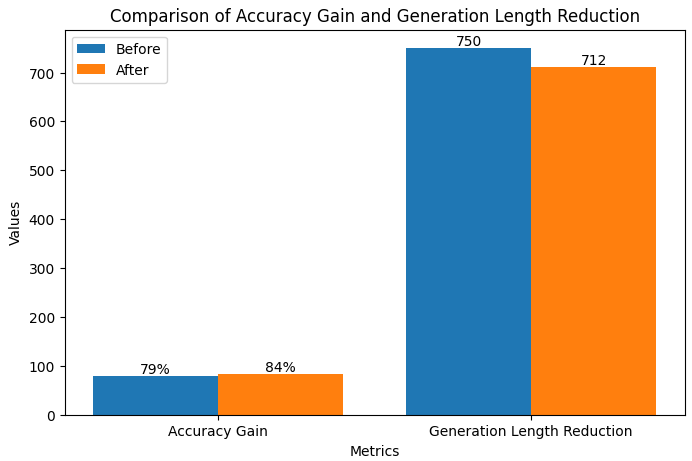}
    \caption{Topological Tuning Results Overall: Improvements in overall accuracy and reduction in generated length are observed from topo-tuned model.}
    \label{fig:fig3}
\end{figure}

To evaluate performance, we test our finetuned model on an out-of-sample test set. Results for Topological Tuning are shown in Figure~\ref{fig:fig3}. The observed \textbf{+5\%} accuracy improvement highlights the benefits of post training with diversely structured, high-quality data, and demonstrates the effectiveness of the TAG mechanism in generating, annotating, and selecting relevant examples. This leads to enhanced complex reasoning capabilities, particularly in problem-solving accuracy. Additionally, the \textbf{5\%} reduction in generated token length suggests the potential for achieving higher accuracy with lower inference latency. The underlying cause of this length reduction merits further investigation. 

\begin{figure}[h!]
    \centering
    \includegraphics[width=0.5\textwidth]{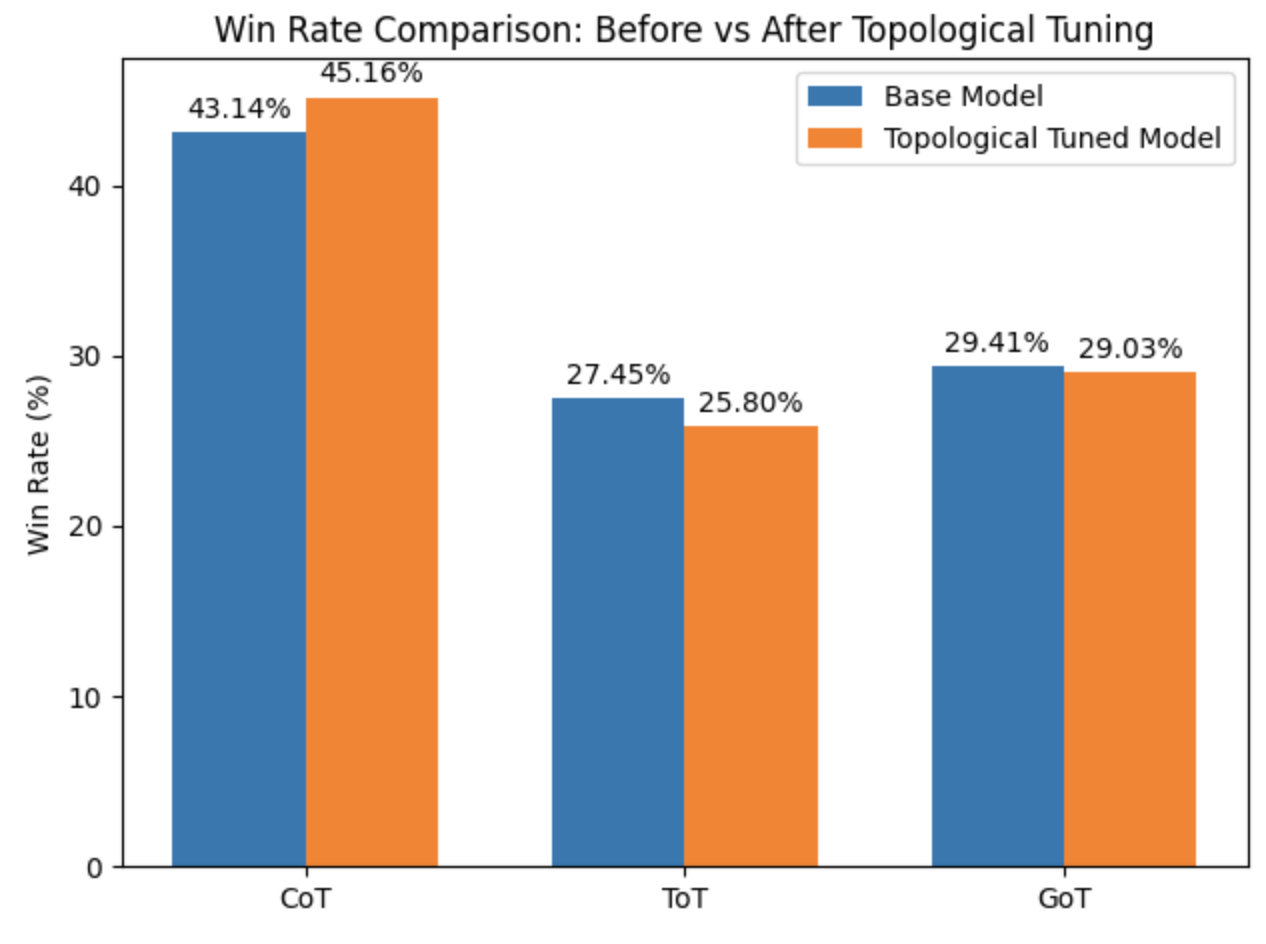}
    \caption{Topo-wise Win Rate Comparison}
    \label{fig:fig2_1}
\end{figure}

\begin{figure}[h!]
    \captionsetup{belowskip=0pt}
    \centering
    \includegraphics[width=0.5\textwidth]{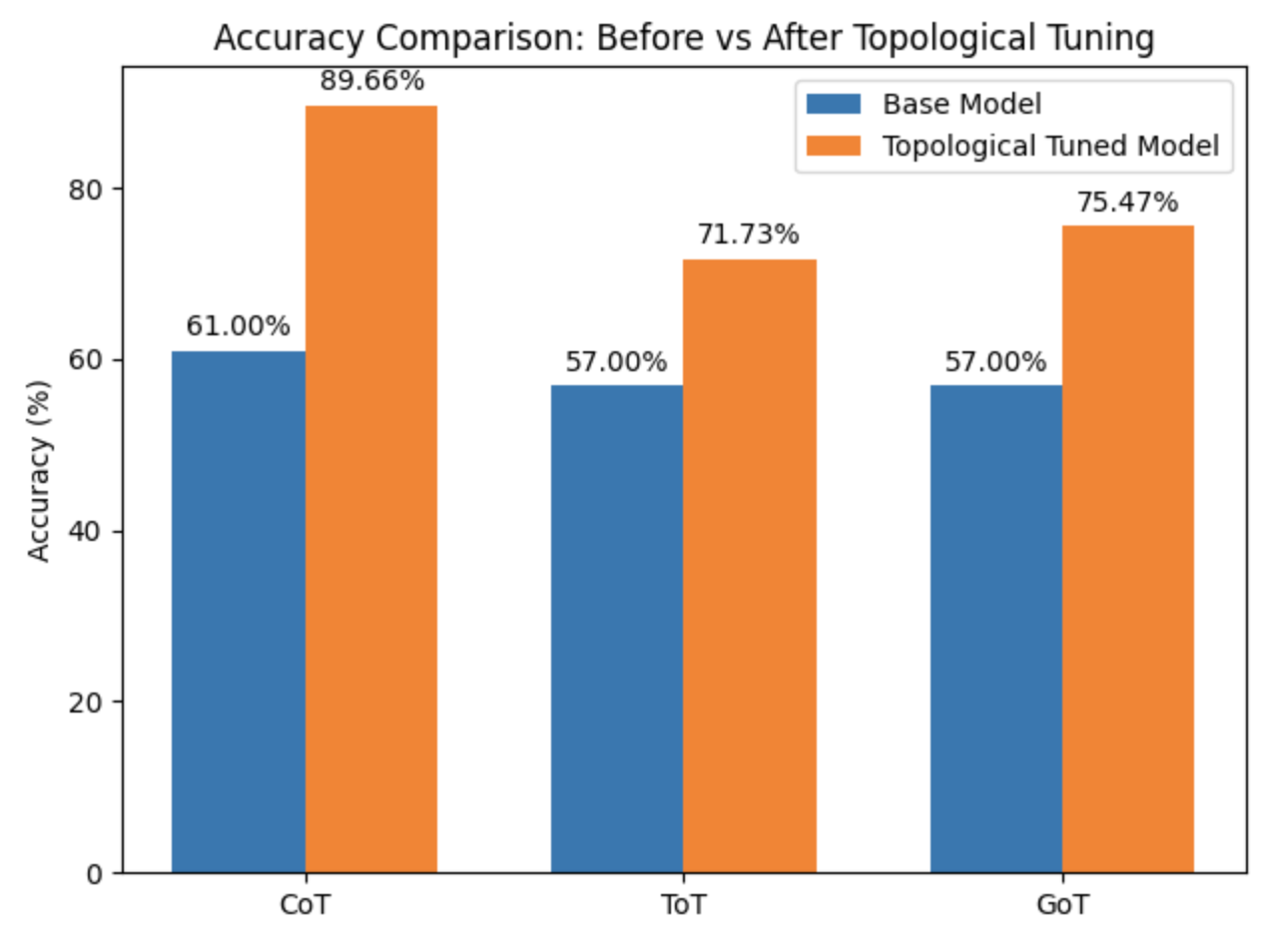}
    \caption{Topo-wise Accuracy Gains}
    \label{fig:fig2_2}
\end{figure}

We further compare our finetuned model with the baseline by explicitly prompting it to reason with all three reasoning topologies. The topology-wise Accuracy and Win Rate before and after Topological Tuning are shown in Figure~\ref{fig:fig2_1} and Figure~\ref{fig:fig2_2} Detailed numbers are show in the Table~\ref{tab:overall_accuracy_comparison} and Table~\ref{tab:win_rate_comparison} in Appendix.

\subsubsection{Ablation Study}  
To rule out that performance gains are solely due to finetuning rather than the multi-topology effect, we conduct an ablation study to assess the additional value provided by the augmented  reasoning topologies. We compare a model finetuned exclusively on CoT data, SFT-Chain—the default behavior of most state-of-the-art reasoning models—against a model finetuned on a mix of all three reasoning topologies. Both models are trained on the same sample size.  

\begin{figure}[h!]
    \centering
    \includegraphics[width=0.5\textwidth]{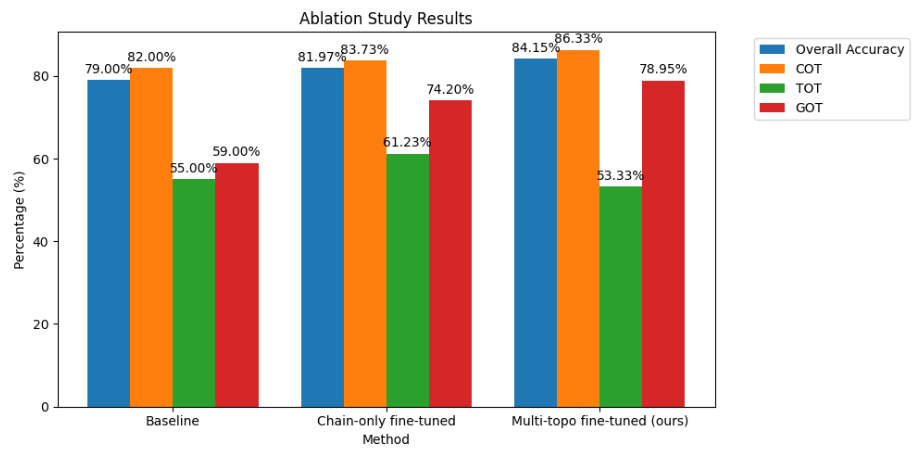}
    \caption{Ablation Study Results: Chain-Only Finetunining vs Topological Tuning}
    \label{fig:ablation}
\end{figure}

Results, shown in Figure~\ref{fig:ablation}, indicate that the multi-topology finetuned model outperforms the SFT-Chain in overall accuracy, CoT accuracy, and GoT accuracy, while exhibiting a slight drop in ToT accuracy. Since these findings confirm that learning from optimal reasoning topologies improves overall accuracy—and given that variations across individual reasoning topologies are expected—the minor decline in ToT performance is acceptable and does not invalidate our main hypothesis. More detailed numbers are in the Table~\ref{tab:accuracy_results_ablation}.

\textbf{Observation 1: Overall Accuracy}  
\vspace{-0.3cm}
\begin{itemize}   
    \item Finetuning solely on CoT does improve accuracy.  
    \item Incorporating ToT and GoT data further enhances performance, demonstrating the added value of diverse topological tuning, eliminating the likelihood that the gain reported in Section~\ref{sec:topo_fintune} is not merely a result of post-training on relevant data, independent of compositional reasoning topologies. Instead, the results demonstrate a robust improvement driven by diverse topological reasoning augmentation. 
\end{itemize}  

\textbf{Observation 2: Topo-wise Accuracy}\newline  
The ablation study demonstrates a \textit{synergistic effect}: Chain-only finetuning boosts GoT accuracy, while mixed training improves CoT. This suggests cross-topology benefits and opens a promising direction for enhancing LLM reasoning.

Results shown in Figure~\ref{fig:fig2_1},  Figure~\ref{fig:fig2_2}, and Figure~\ref{fig:ablation} collectively demonstrate the effectiveness of Topological Tuning.  

\subsection{Impact of Topological Rewarding and Hybrid Scaling}\label{sec:inf_scale}
We next assess \emph{Topological Rewarding} (inference scaling only) and \emph{Hybrid Scaling} (combining training- and inference-scale optimizations). For Topological Rewarding, we generate multi-topology reasoning based responses using a non-finetuned Qwen2-VL-7B-Instruct model, and then apply our multi-task Topological Reward Model (M-TRM) to select the optimal topology and the best final answer. 

\begin{figure}[h!]
    \centering
    \includegraphics[width=0.5\textwidth]{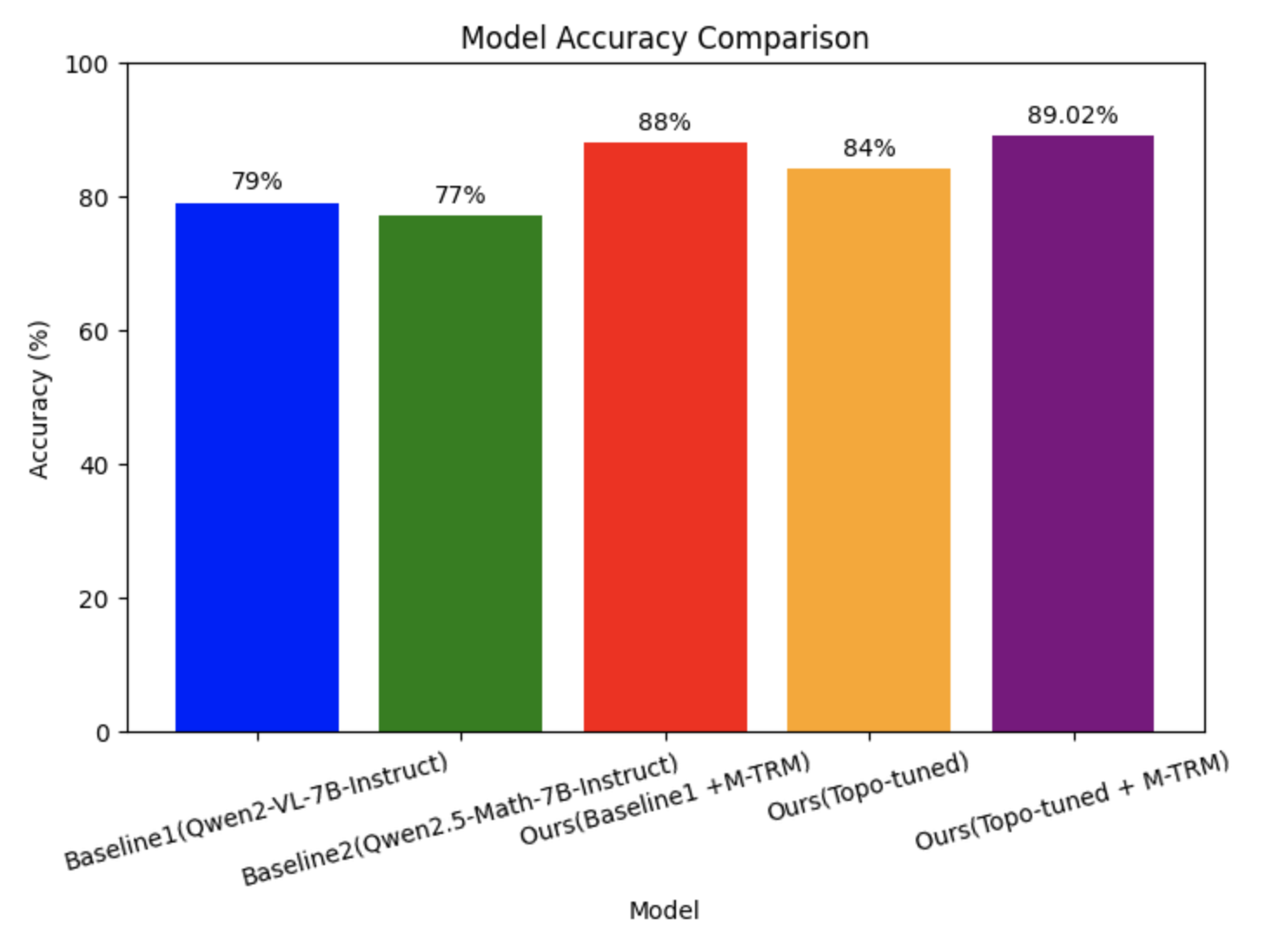}
    \caption{Performance of two Baselines and three our methods: Hybrid Scaling achieves the highest accuracy}
    \label{fig:fig6}
\end{figure}

Hybrid Scaling adds Topological Rewarding atop a topologically tuned model using the same selection process. Figure~\ref{fig:fig6} shows that Topological Tuning alone boosts accuracy by \textbf{+5\%}, Topological Rewarding by \textbf{+9\%}, and Hybrid Scaling by \textbf{+10.02\%}. Detailed results are in Table~\ref{tab:accuracy_latency}.

\subsection{Discussion and Limitations}\label{sec:discuss}
As shown in Table~\ref{tab:accuracy_latency}, Topological Tuning, Topological Rewarding, and Hybrid Scaling each boost performance but incur different inference latencies. Topological Tuning is especially beneficial in latency-sensitive contexts, as it shortens responses for complex tasks, potentially mitigating overthinking through exposure to both ``winning'' and ``losing'' topologies. Further investigation of this effect remains future work.

ToT and GoT topologies offer non-Euclidean structures that address diverse problem complexities, and our framework automatically selects the most effective approach. We considered but dismissed graph traversal for chain construction, finding minimal gains-due to its high similarity to CoT, relative to its overhead (Section~\ref{sec:inf_scale}).

Finally, our results focus on in-domain datasets; broader generalization requires advanced post training methods such as Reinforcement Learning \citep{chu2025sftmemorizesrlgeneralizes}. We are exploring this direction to demonstrate its efficacy on out-of-domain tasks.

\section{Conclusion}
We have presented \textbf{SOLAR}, a paradigm shift in LLM reasoning that learns to adaptively choose among Chain-of-Thought, Tree-of-Thought, or Graph-of-Thought reasoning strategies. By unifying post training and inference-scale optimizations, SOLAR not only generates effective policies but also refines candidate solutions through a competitive selection process, substantially improving performance on both MATH and GSM8K.  

Our experiments validate the effectiveness of \emph{Topological-Annotation-Generation} (TAG) and curriculum learning-based \emph{Topological Scaling} in enhancing adaptive reasoning beyond the conventional chain-of-thought. Notably, we observe a reduction in response length for complex tasks—an effect we refer to as \emph{resilience to overthinking}—demonstrating SOLAR's ability to streamline outputs without compromising accuracy.

Our research opens several promising avenues for further exploration. Two key questions arise: 
How can we further optimize the \textit{synergistic effect} between reasoning structures and scaling laws? 
What internal factors drive Vision-Language Models (e.g., Qwen2VL) to increase the likelihood of non-default reasoning topologies, and how such drivers are related to its development life cycle (e.g.from pretraining to post training)?
Additionally, what underlying principles account for our approach’s anti-overthinking behavior? 
We are also integrating RL-based algorithms to enhance the generalization capabilities of our method.
Addressing these questions not only deepens our understanding of LLM cognition but also unlocks new frontiers in adaptive reasoning architectures, paving the way for more scalable, efficient, and ethical AI systems.

\nocite{langley00}

\bibliography{example_paper}
\bibliographystyle{icml2025}

\newpage
\appendix
\onecolumn
\begin{table}[H]
    \centering
    \renewcommand{\arraystretch}{0.9} 
    \setlength{\tabcolsep}{3pt} 
    \caption{Accuracy and Win Rate: Non-fintuned Pretrained-Model}
    \label{tab:accuracy_winrate}
    \begin{tabular}{lccc}
        \toprule
        \textbf{Model} & \textbf{CoT} & \textbf{ToT} & \textbf{GoT} \\
        \midrule
        \multicolumn{4}{c}{\textbf{Accuracy (\%)}} \\
        \midrule
        Large Reasoning Model & 82.86 & 78.19 & 76.75 \\
        Qwen2-VL-7B-Instruct & 69.32 & 64.82 & 67.18 \\
        \midrule
        \multicolumn{4}{c}{\textbf{Win Rate (\%)}} \\
        \midrule
        Large Reasoning Model & 35.86 & 32.58 & 31.54 \\
        Qwen2-VL-7B-Instruct & 28.96 & 23.88 & 26.06 \\
        \bottomrule
    \end{tabular}
\end{table}

\begin{table}[H]
    \centering
    \caption{Ablation Study Accuracy}
    \label{tab:accuracy_results_ablation}
    \begin{tabular}{lcccccc}
        \toprule
        \textbf{Method} & \textbf{Overall} & \textbf{CoT} & \textbf{ToT} & \textbf{GoT} \\
        \midrule
        Baseline (Qwen2-VL-7B-Instruct) & 79\% & 82\% & 55\% & 59\% \\
        Chain-only finetuned & 81.97\% & 83.73\% & 61.23\% & 74.20\% \\
        Multi-topo finetuned (ours) & 84.15\% & 86.33\% & 53.33\% & 78.95\% \\
        \bottomrule
    \end{tabular}
\end{table}

\begin{table}[h]
    \centering
    \renewcommand{\arraystretch}{0.9} 
    \setlength{\tabcolsep}{4pt} 
    \caption{Overall Accuracy Comparison: Baseline Model vs. Topological Tuned Model}
    \label{tab:overall_accuracy_comparison}
    \small 
    \begin{tabular}{lccc}
        \toprule
        \textbf{Model} & \textbf{CoT} & \textbf{ToT} & \textbf{GoT} \\
        \midrule
        Baseline Model (Qwen2-VL-7B-Instruct) & 61.00\% & 57.00\% & 57.00\% \\
        Topological Tuned Model & 89.66\% & 71.73\% & 75.47\% \\
        \bottomrule
    \end{tabular}
\end{table}

\begin{table}[H]
    \centering
    \renewcommand{\arraystretch}{0.9} 
    \setlength{\tabcolsep}{4pt} 
    \caption{Win Rate Comparison: Baseline Model vs. Topological Tuned Model}
    \label{tab:win_rate_comparison}
    \small 
    \begin{tabular}{lccc}
        \toprule
        \textbf{Model} & \textbf{CoT} & \textbf{ToT} & \textbf{GoT} \\
        \midrule
        Baseline Model (Qwen2-VL-7B-Instruct) & 43.14\% & 27.45\% & 29.41\% \\
        Topological Tuned Model & 45.16\% & 25.80\% & 29.03\% \\
        \bottomrule
    \end{tabular}
\end{table}

\begin{table}[H]
    \centering
    \caption{Accuracy for Topological Scaling Comparison}
    \label{tab:accuracy_latency}
    \begin{tabular}{lcccccc}
        \toprule
        \textbf{Method} & \textbf{Overall Accuracy} & \textbf{Test Latency} \\
        \midrule
        Baseline1 (Qwen2-VL-7B-Instruct) & 79\% & Medium \\
        Baseline2 (Qwen2.5-Math-7B-Instruct) & 77\% & Medium \\
        Topo-Rewarding (ours) & 88\% & High  \\
        Topo-Tuning (ours) & 84\% & Low \\
        Hybrid-Scaling (ours) & 89.02\% & Medium to High \\
        \bottomrule
    \end{tabular}
\end{table}

\end{document}